\documentclass[hidelinks]{article}

\usepackage[preprint]{neurips_2023}

\usepackage{booktabs, graphicx, subcaption, pgfplots}
\usepackage[inline]{enumitem}
\usepackage[utf8]{inputenc} 
\usepackage[T1]{fontenc}    
\usepackage{hyperref}       
\usepackage{url}            
\usepackage{booktabs}       
\usepackage{amsfonts}       
\usepackage{nicefrac}       
\usepackage{microtype}      
\usepackage{xcolor}         

\usepackage{natbib}

\title{Do Generative Large Language Models need billions of parameters?}

\author{
    Sia Gholami \\
    The Institute of Electrical and Electronics Engineers, Member IEEE \\
    \texttt{gholami@ieee.org} \\
    \And 
    Marwan Omar \\
    Illinois Institute of Technology \\
    \texttt{momar3@iit.edu} \\
}

\begin{document}

\maketitle

\begin{abstract}
This paper presents novel systems and methodologies for the development of efficient large language models (LLMs). It explores the trade-offs between model size, performance, and computational resources, with the aim of maximizing the efficiency of these AI systems. The research explores novel methods that allow different parts of the model to share parameters, reducing the total number of unique parameters required. This approach ensures that the model remains compact without sacrificing its ability to learn and represent complex language structures. This study provides valuable insights and tools for creating more efficient and effective LLMs, contributing to a more sustainable and accessible future for AI language modeling.
\end{abstract}

\section{Introduction}

Large language models (LLMs) are a type of artificial intelligence (AI) model designed to understand and generate human-like text based on the context provided to them. These models have become increasingly popular in recent years, with the most notable examples being OpenAI's GPT series~\citep{radford2019language}, including the latest version, GPT-4. LLMs have shown significant improvements over previous models in various natural language processing (NLP) tasks and have numerous practical applications in diverse domains. LLMs are a subcategory of deep learning models that focus on processing and generating text data. They are built using transformer architectures, which were first introduced by ~\citep{vaswani2017attention}. and have since revolutionized the field of NLP. Transformers are characterized by their self-attention mechanisms, which enable them to process and generate text sequences in parallel rather than sequentially, as was the case with previous models like recurrent neural networks (RNNs) and long short-term memory (LSTM) networks.

Large language models are trained on vast amounts of text data, often sourced from the internet, to learn patterns, structures, and relationships within the text. This allows them to understand and generate human-like text in a variety of contexts~\citep{shoeybi2019megatron}. The size of these models is typically measured in terms of parameters, with recent models like GPT-3 and GPT-4 boasting billions of parameters~\citep{fedus2022switch}.

Large language models have numerous practical applications across a wide range of domains, including but not limited to:

\begin{enumerate}
    \item Text Generation and Completion: LLMs can generate human-like text based on a given prompt, making them valuable tools for tasks like content creation, email drafting, and code generation \citep{brown2020language}.

    \item Machine Translation: LLMs have demonstrated impressive performance in translating text between languages, rivaling dedicated machine translation models \citep{lample2019cross}.

    \item Summarization: LLMs can generate concise summaries of longer text passages, making them useful in areas like news aggregation, research paper summarization, and extracting key information from documents \citep{liu2019text}.

    \item Question Answering: LLMs can understand and answer questions based on a given context, making them effective tools for applications like virtual assistants, customer support, and knowledge extraction \citep{rajpurkar2016squad}.

    \item Sentiment Analysis: LLMs can classify text based on sentiment, enabling use cases in market research, social media monitoring, and product review analysis \citep{zhang2018deep}.

    \item Conversational AI: LLMs can generate contextually appropriate responses in a dialogue, making them useful for building chatbots and voice assistants \citep{adiwardana2020towards}.
\end{enumerate}

Despite their impressive capabilities, large language models also present several challenges related to their development, deployment, and ethical considerations:

\begin{itemize}
    \item Computational and Energy Requirements: Training LLMs requires vast computational resources, making the process expensive and energy-intensive\citep{strubell2019energy}. This limits accessibility to such models for researchers and organizations with limited resources.

    \item Data Bias: Since LLMs are trained on large datasets sourced from the internet, they may inadvertently learn and propagate biases present in the data  \citep{bender2021dangers}. This can lead to biased outputs, negatively impacting certain user groups or perpetuating harmful stereotypes.

    \item Model Controllability: Controlling the outputs of LLMs can be challenging due to their complex and highly interconnected nature. This can generate inappropriate, harmful, or offensive content \citep{radford2019language}.

    \item Intellectual Property Concerns: LLMs can generate text that resembles human-written content, which raises questions about intellectual property rights and potential copyright infringement \citep{lipton2018troubling}.

    \item Privacy Issues: As LLMs are trained on vast amounts of data, there is a possibility that they may unintentionally memorize and reveal sensitive information, presenting privacy concerns \citep{carlini2019secret}.
\end{itemize}

The primary goal of this study is to explore, investigate and propose new approaches to create efficient LLMs. This objective is important for several reasons: 

\begin{itemize}
    \item Accessibility: Efficient LLMs require fewer computational resources for training and inference, making them more accessible to researchers, developers, and organizations with limited resources. This democratizes access to state-of-the-art natural language processing (NLP) technology, promoting innovation and reducing the gap between well-funded organizations and smaller players in the field.

    \item Energy consumption: Training and deploying LLMs can be energy-intensive due to their size and complexity (Strubell et al., 2019). Developing more efficient models can significantly reduce energy consumption and the associated environmental impact, contributing to more sustainable AI development practices.

    \item Cost reduction: Efficient LLMs can lower the financial costs associated with training, deployment, and maintenance. This allows organizations to allocate resources more effectively, potentially accelerating the development of new applications and services that rely on advanced NLP capabilities.

    \item Latency and real-time applications: Efficient LLMs can provide faster response times during inference, which is particularly important for real-time applications, such as conversational AI, virtual assistants, and other interactive systems that demand low latency.

    \item Edge computing: As more AI applications are being deployed on edge devices, such as smartphones and IoT devices, it is crucial to develop efficient LLMs that can operate within the constraints of limited computing power, memory, and energy resources. Efficient LLMs can enable sophisticated NLP capabilities on edge devices, expanding their potential use cases and improving user experiences.

    \item Scalability: Developing efficient LLMs allows for better scalability in terms of the number of users and applications supported by a given model. This is particularly relevant for cloud-based AI services that need to serve a large number of clients simultaneously, where resource efficiency directly translates into cost savings and improved performance.

    \item Encouraging research: Making LLMs more efficient can stimulate further research in model optimization, compression, and resource-conscious training techniques. This can lead to the discovery of novel methods that improve LLMs and other types of deep learning models, benefiting the broader AI research community.
\end{itemize}

Parameter sharing can indeed help make Transformer models more efficient. Transformers are known for their large number of parameters, which can lead to high computational and memory requirements. By sharing parameters across different parts of the model, you can reduce the total number of unique parameters, leading to a smaller and more efficient model.

Using parameter sharing in Transformers can be beneficial for several reasons:

\begin{itemize}
    \item Reduced model size: Transformers typically have a large number of parameters, which can result in substantial memory and storage requirements. Parameter sharing helps reduce the overall number of unique parameters, leading to a more compact model that is easier to store and deploy.

    \item Faster training and inference: With fewer parameters to learn and process, training and inference times can be reduced, leading to more efficient and faster models. This can be especially important when deploying models on resource-constrained devices or in situations where low-latency responses are critical.

    \item Improved generalization: Sharing parameters across different parts of the model can encourage the learning of more general and robust features. This can potentially lead to better generalization to unseen data or tasks, improving the model's overall performance.

    \item Easier optimization: With fewer parameters to optimize, the model's optimization landscape may become smoother and easier to navigate. This can lead to faster convergence during training and potentially better final model performance.

    \item Regularization effect: Parameter sharing can have a regularization effect by implicitly constraining the model's capacity. This can help prevent overfitting, especially when training data is limited.

    \item Transfer learning and multi-task learning: Parameter sharing can be used to share representations between different tasks or modalities, promoting the learning of shared features and potentially improving performance on multiple tasks simultaneously.
\end{itemize}

However, it's important to strike a balance between the benefits of parameter sharing and the potential reduction in model performance. Excessive parameter sharing can lead to a loss of expressive power, resulting in decreased performance on the given task. It is crucial to carefully consider the trade-offs and determine the appropriate level of parameter sharing that maximizes efficiency while maintaining the desired performance.

\section{Related Works}
Natural Language Processing (NLP) has been a major area of research in Artificial Intelligence and Machine Learning
since the early days of computer science~\citep{voorhees1999trec, moldovan2000structure, brill2002analysis, ferrucci2010building, gholami2021zero, gholami2022you, gholami2022create, gholami2022alexa, gholami2022flight, brand2022text}. Parameter sharing has been employed in several transformer models to improve efficiency and reduce the number of parameters. Some notable examples include:

Universal Transformers~ \citep{dehghani2018universal}: Universal Transformers extend the original transformer architecture by sharing parameters across layers. Instead of having separate parameters for each layer, the same set of weights is applied recursively for a fixed number of steps, updating the hidden states at each step. This approach substantially reduces the number of parameters while maintaining competitive performance on various NLP tasks.

ALBERT~ \citep{lan2019albert}: ALBERT (A Lite BERT) shares all parameters across layers, significantly reducing model size compared to the original BERT. By doing so, ALBERT achieves comparable performance to larger models with fewer parameters, making it more efficient and easier to deploy.

LayerDrop~ \citep{fan2019reducing}: LayerDrop is a technique that randomly drops a subset of layers during training, effectively sharing parameters between different "depths'' of the transformer. This approach can help improve efficiency and model performance without significantly increasing the number of parameters.

Tied Transformers~ \citep{lample2017unsupervised}: Tied Transformers use parameter sharing between the encoder and the decoder, sharing not only the layer parameters but also the token embeddings and positional embeddings. This approach reduces the number of parameters and improves the model's efficiency, particularly in unsupervised machine translation tasks.

Depthwise Separable Transformers~ \citep{kitaev2020reformer}: Depthwise Separable Transformers employ parameter sharing across different attention heads within the same layer. This method reduces the number of parameters while preserving the benefits of multi-head attention, leading to a more efficient model.

Longformer~ \citep{beltagy2020longformer}: Longformer utilizes a combination of local and global attention mechanisms, sharing parameters between local and global attention heads. This approach allows the model to process long documents efficiently without a significant increase in the number of parameters.

Big Bird~ \citep{zaheer2020big}: Big Bird employs a combination of sparse attention mechanisms and parameter sharing across different attention heads to create an efficient transformer capable of handling longer sequences. This approach reduces both the computational complexity and the number of parameters compared to the original transformer architecture.

Switch Transformers~ \citep{fedus2022switch}: Switch Transformers introduce a novel parameter-sharing technique called "mixture of experts," in which different parts of the model specialize in different aspects of the data. By sharing parameters across experts and selectively activating only a subset of them during inference, Switch Transformers can scale to trillion-parameter models while maintaining high efficiency.

These examples demonstrate the effectiveness of parameter sharing in creating more efficient transformers while maintaining competitive performance on various NLP tasks.

\section{Approach}
Transformers were introduced by Vaswani et al.\citep{vaswani2017attention} as a novel architecture for NLP tasks, relying on self-attention mechanisms to capture long-range dependencies in the input data. They consist of a stack of identical layers, each containing a multi-head self-attention mechanism and position-wise feed-forward networks. The self-attention mechanism allows the model to weigh the importance of different tokens in the input sequence relative to each other, enabling it to learn complex patterns and dependencies effectively.

Despite their impressive performance, transformers have high computational and memory requirements. These demands stem from the quadratic complexity of the self-attention mechanism, which requires calculating attention scores for each token pair in the input sequence. Additionally, transformers often have millions of parameters, making them resource-intensive to store and update during training and inference \citep{yun2020n}.

Parameter sharing is a technique used to reduce the number of model parameters by reusing them across different components of the model. This approach has been successfully applied in convolutional neural networks (CNNs) for image recognition tasks, where the same set of weights is used across different spatial locations, significantly reducing the number of parameters.

Parameter sharing offers several benefits in building efficient transformers:

\begin{itemize}
    \item Reduced Model Size: By reusing parameters across different layers or components, parameter sharing substantially decreases the number of unique parameters required to represent the model. This reduction in model size makes it easier to deploy transformers on resource-constrained devices, where memory is often limited.

    \item Faster Training and Inference: With fewer parameters to update during training, the model can be trained more quickly, leading to faster convergence. This speed-up can be particularly beneficial when training large models or working with large-scale datasets. Moreover, during inference, a smaller model requires fewer computations, which translates to faster prediction times.

    \item Regularization Effect: Parameter sharing can act as a form of implicit regularization, constraining the model's capacity and reducing overfitting. By reusing parameters, the model is forced to learn shared representations, which can lead to better generalization performance on unseen data.
\end{itemize}

Several parameter-sharing \citep{radford2018improving} techniques have been proposed and explored to improve the efficiency of transformers. We discuss some of the most prominent approaches: 
\begin{itemize}
    \item Universal Transformers: Universal Transformers \citep{dehghani2018universal} extend the original transformer architecture by sharing parameters across layers. Instead of having separate parameters for each layer, Universal Transformers apply the same set of weights recursively for a fixed number of steps, updating the hidden states at each step. This approach significantly reduces the number of parameters while maintaining competitive performance on various NLP tasks.

    \item Layer-wise Parameter Sharing: Layer-wise parameter sharing involves reusing parameters across multiple layers in the transformer. For example, ALBERT \citep{lan2019albert} shares all parameters across layers, leading to a substantial reduction in model size. By doing so, ALBERT achieves comparable performance to larger models with fewer parameters, making it more efficient and easier to deploy.

    \item Head-wise Parameter Sharing: In transformers, the multi-head self-attention mechanism computes multiple attention scores for each token pair. Head-wise parameter sharing involves reusing the same set of weights across different attention heads within the same layer. This approach can help reduce the number of parameters while preserving the benefits of multi-head attention.
    
    \item Parameter Sharing in Feed-Forward Networks: Transformers also include position-wise feed-forward networks in each layer, which can be another source of parameter reduction. Sharing parameters across the feed-forward networks of different layers can help achieve a more compact model while maintaining its capacity to learn complex patterns.

    \item Pre-trained Models and Fine-tuning: Parameter sharing can also be employed through the use of pre-trained transformer models, such as BERT \citep{devlin2018bert} and GPT \citep{radford2018improving}. By fine-tuning these pre-trained models on specific tasks, it is possible to leverage the shared knowledge encoded in the model's parameters and achieve strong performance with fewer training samples and reduced training time.
\end{itemize}

Parameter sharing is not without its drawbacks and challenges. Some of the key issues include:

\begin{itemize}
    \item Reduced Expressive Power: Sharing parameters can limit the model's expressive power, as it is forced to learn shared representations. In some cases, this reduction in capacity can lead to a degradation in performance, especially when the shared parameters are not suitable for the task at hand.

    \item Hyperparameter Tuning: Parameter sharing introduces additional hyperparameters, such as the number of shared layers or the extent of sharing within a layer. Finding the optimal configuration of these hyperparameters can be challenging and may require extensive experimentation and search.

    \item Task-Specific Considerations: Not all tasks may benefit from parameter sharing, and the optimal degree of sharing can vary depending on the nature of the task. For example, tasks that require modeling fine-grained dependencies or learning highly specialized representations may suffer from performance degradation with increased parameter sharing.
\end{itemize}

Parameter sharing has emerged as a promising approach to building efficient transformers, offering benefits such as reduced model size, faster training and inference, and implicit regularization. Various parameter sharing techniques have been proposed and explored, including layer-wise, head-wise, and feed-forward network sharing, as well as leveraging pre-trained models.


To enhance the Transformer model's efficiency and efficacy for self-supervised learning of language representations. The suggested technique does this by lowering the number of model parameters while maintaining or enhancing performance. The proposed technique uses two key strategies: parameter sharing and factorized embedding parameterization across layers, which results in a smaller model with fewer parameters. I call the transformer created by these techniques Generative Pretrained Transformer Efficio (GPT-Efficio). 

\subsubsection{Embedding layer factorization:}
In this approach, the original embedding layer is factorized into two matrices, which reduces the total number of parameters in the model. This is particularly useful for large-scale models like BERT and GPT, where the embedding layer can consume a significant amount of memory due to the large vocabulary size. The factorization works by separating the original embedding layer into two smaller layers:

\begin{enumerate}
    \item A word-piece embedding matrix that maps the input tokens into a lower-dimensional space ($EmbeddingSize$).
    \item A projection matrix that maps the lower-dimensional embeddings into the model's hidden state dimension ($HiddenSize$).
\end{enumerate}

The original embedding matrix of the decoder transformer has a total of $VocabSize\times HiddenSize$ parameters. To get better performance, NLP models tend to use large vocabulary size and this layer can be billions of parameters \citep{keskar2016large}. 

In this approach, I break up the embedding matrix into two smaller matrices, this will bring down the number of parameters to $VocabSize\times EmbeddingSize + EmbeddingSize\times HiddenSize$. This technique is more effective when the $EmbeddingSize$ is significantly smaller than the $HiddenSize$.

However, this kind of factorization can lead to a loss of information. Mapping the tokens to a smaller dimension might not capture all the nuances of the input data. In turn, this might affect the performance of the model, causing it to make more errors or have a lower overall accuracy. 

\subsubsection{Parameter sharing across layers:}
In my approach GPT-Efficio, the adoption of parameter sharing strategies is prevalent to mitigate the escalating complexity arising from an abundance of parameters. By sharing parameters, the model size can be substantially reduced, consequently enhancing generalization by introducing repetition in the underlying structure of the model.

In this context, I outline several methodologies for sharing parameters in GPT-Efficio:

\begin{enumerate}
    \item Layer-wise Parameter Sharing: This technique involves sharing the same set of parameters across all transformer layers, which significantly diminishes the model size. This can be practically implemented, for instance, in PyTorch by defining a single transformer layer that is repeatedly employed in a loop for all layers. However, this method might potentially curtail the model's expressiveness, given that identical transformations are propagated at each layer.
    \item Sub-layer Parameter Sharing: Here, the same parameters are shared between two principal components of each transformer layer, namely the self-attention and feed-forward sub-layers. This method further contributes to the reduction of model complexity.
    \item Sub-layer Parameter Sharing in Groups: In this approach, the same parameters are shared between two principal components of each transformer layer, namely the self-attention and feed-forward sub-layers in groups. For example if the group number is 1, all the parameters are shared across all layers, if the group number is 2, the parameters will be shared between half of the layers and so on. The number of groups is a hyperparameter that should be optimized for the downstream tasks.  
\end{enumerate}

While the benefits of parameter sharing are evident in terms of model size reduction and potential generalization improvements, it is important to consider that these benefits may not uniformly translate across all problems. The risk of limiting the model's capacity and expressiveness is a noteworthy drawback that necessitates careful consideration of the specific use case when deciding on the adoption of parameter sharing techniques.

\section{Experiments}

In this chapter we present the results of each of our approaches in the context of language modeling (i.e. completion tasks) and question answering.

\subsection{Results}

The implementation of parameter sharing strategies in GPT-Efficio, yields an array of outcomes that significantly affect the performance and computational efficiency of these models.

Positive outcomes include:

\begin{enumerate}
    \item Diminished Model Complexity: Parameter sharing can drastically reduce the count of unique trainable parameters, thereby decreasing the overall complexity of the model. This benefit is particularly advantageous in scenarios where computational resources or memory availability are constrained.
    \item Expedited Training Process: Given fewer unique parameters, the training process can be substantially accelerated. This advantage becomes increasingly significant when dealing with large-scale datasets or in scenarios where time efficiency is a key consideration.
    \item Enhanced Generalization: The adoption of parameter sharing could prompt the model to learn more generalized features that apply across various parts of the input. For instance, sharing weights across all positions in a sequence may force the model to learn position-independent features. Similarly, sharing parameters across different layers or sub-layers can promote the extraction of features with broader applicability \citep{bengio1994learning}.
    \item Mitigation of Overfitting: By reducing the number of parameters, parameter sharing strategies can contribute to the prevention of overfitting, especially in situations where the training data available is relatively scarce.
\end{enumerate}

Conversely, there exist drawbacks that merit consideration:

\begin{enumerate}
    \item Limited Model Expressivity: Although parameter sharing can contribute to complexity reduction and improved generalization, it could concurrently constrain the model's capacity. If identical parameters are applied across different layers or sub-layers, the model might be inhibited in its ability to learn distinct representations at different depths of the network, potentially compromising its performance on complex tasks.
    \item Risk of Sub-optimal Solutions: There may be instances where the optimal solution necessitates different parameters in different components of the model. In such cases, parameter sharing could potentially culminate in sub-optimal solutions.
    \item Dependence on Task and Dataset Specificity: The effectiveness of parameter sharing can be heavily influenced by the specific task and dataset in use. It is not guaranteed to invariably lead to performance improvements and could, in certain cases, negatively impact the model's performance.
\end{enumerate}

\begin{table}[!htbp]
\centering
\small
\caption{Performance of parameter sharing approach on completion tasks}\label{param-lm}
\begin{tabular}{p{0.2\linewidth} p{0.1\linewidth} p{0.15\linewidth} p{0.15\linewidth} p{0.12\linewidth} p{0.12\linewidth}}
\toprule
\textbf{Model} & \textbf{$n_{params}$} & \textbf{LAMBADA (acc)} & \textbf{LAMBADA (ppl)} & \textbf{StoryCloze (acc)} & \textbf{HellaSwag (acc)} \\ 
\midrule
GPT-3 Zero-Shot & 175B & 76.2 & 3.00 & 83.2 & 78.9   \\ 
GPT-3 One-Shot & 175B & 72.5 & 3.35 & 84.7 & 78.1   \\ 
GPT-3 Few-Shot & 175B & 86.4 & 1.92 & 87.7 & 79.3   \\ 
GPT-Efficio & 950M & 67.1 & 9.2 & 80.5 & 72.6 \\
\bottomrule
\end{tabular}
\end{table}

Table~\ref{param-lm} demonstrates the GPT-Efficio performance in comparison with GPT-3. 

\pgfplotsset{width=0.9\textwidth}
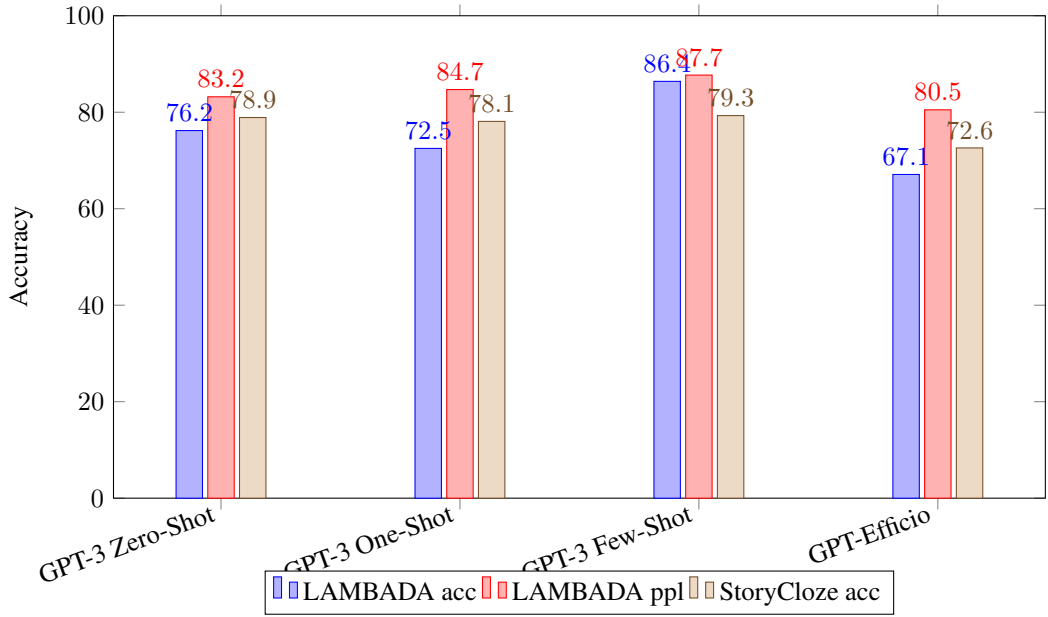
\begin{figure}[htbp]
\centering
\begin{tikzpicture}
    \begin{axis}[
        width=\linewidth,
        height=8cm,
        ybar, ymin=0, ymax=100, 
        enlarge x limits=0.15,
        legend style={at={(0.5,-0.15)},
        anchor=north,legend columns=-1},
        ylabel={Accuracy},
        symbolic x coords={GPT-3 Zero-Shot, GPT-3 One-Shot, GPT-3 Few-Shot, GPT-Efficio},
        xtick=data,
        nodes near coords,
        nodes near coords align={vertical},
        x tick label style={rotate=20,anchor=east},
        ]
        \addplot coordinates {(GPT-3 Zero-Shot,76.2) (GPT-3 One-Shot,72.5) (GPT-3 Few-Shot,86.4) (GPT-Efficio,67.1)};
        \addplot coordinates {(GPT-3 Zero-Shot,83.2) (GPT-3 One-Shot,84.7) (GPT-3 Few-Shot,87.7) (GPT-Efficio,80.5)};
        \addplot coordinates {(GPT-3 Zero-Shot,78.9) (GPT-3 One-Shot,78.1) (GPT-3 Few-Shot,79.3) (GPT-Efficio,72.6)};
        \legend{LAMBADA acc, LAMBADA ppl, StoryCloze acc, HellaSwag acc}
    \end{axis}
\end{tikzpicture}
\caption{Performance of parameter sharing approach on completion tasks}
\label{fig:param-lm}
\end{figure}

\begin{table}[!htbp]
\centering
\small
\caption{Performance of parameter sharing approach on QA tasks}\label{param-qa}
\begin{tabular}{p{0.2\linewidth} p{0.1\linewidth} p{0.15\linewidth} p{0.15\linewidth} p{0.12\linewidth}}
\toprule
\textbf{Model} & \textbf{$n_{params}$} & \textbf{NQ} & \textbf{WebQ} & \textbf{TriviaQA}\\ 
\midrule
GPT-3 Zero-Shot & 175B & 14.6 & 14.4 & 64.3   \\ 
GPT-3 One-Shot & 175B & 23.0 & 25.3 & 68.0   \\ 
GPT-3 Few-Shot & 175B & 29.9 & 41.5 & 71.2   \\ 
GPT-Efficio & 950M & 27.5 & 40.6 & 69.2 \\
\bottomrule
\end{tabular}
\end{table}

Table~\ref{param-qa} shows the GPT-Efficio performance in comparison with GPT-3.

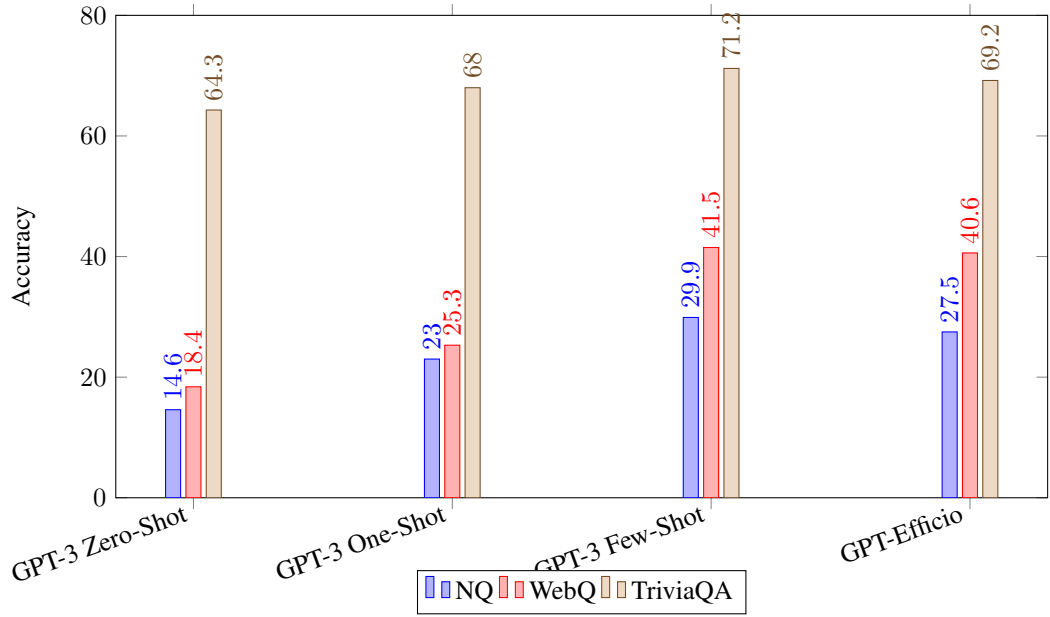
\begin{figure}[!htbp]
\centering
\begin{tikzpicture}
\begin{axis}[
    width=\linewidth,
    height=8cm,
    ybar, ymin=0, ymax=80,
    bar width=0.2cm,
    ylabel={Accuracy},
    symbolic x coords={GPT-3 Zero-Shot, GPT-3 One-Shot, GPT-3 Few-Shot, GPT-Efficio},
    xtick=data,
    nodes near coords,
    nodes near coords align={vertical},
    every node near coord/.append style={rotate=90, anchor=west},
    legend style={at={(0.5,-0.15)},
      anchor=north,legend columns=-1},
    x tick label style={rotate=20,anchor=east},
    ]
\addplot coordinates {(GPT-3 Zero-Shot, 14.6) (GPT-3 One-Shot, 23.0) (GPT-3 Few-Shot, 29.9) (GPT-Efficio, 27.5)};
\addplot coordinates {(GPT-3 Zero-Shot, 18.4) (GPT-3 One-Shot, 25.3) (GPT-3 Few-Shot, 41.5) (GPT-Efficio, 40.6)};
\addplot coordinates {(GPT-3 Zero-Shot, 64.3) (GPT-3 One-Shot, 68.0) (GPT-3 Few-Shot, 71.2) (GPT-Efficio, 69.2)};
\legend{NQ, WebQ, TriviaQA}
\end{axis}
\end{tikzpicture}
\caption{Performance of parameter sharing approach on QA tasks}\label{graph-qa}
\end{figure}


\section{Analysis}
This section investigates the effects of each hyperparameter in a transformer model and its influences on the performance:

\begin{enumerate}
    \item Model Size ($d_{model}$): A larger model size typically enables the model to learn more complex representations, but it also increases the risk of overfitting and requires more computational resources. 
    \item Number of Layers ($num_{layers}$): More layers allow the model to learn more complex, hierarchical representations. However, adding too many layers can lead to difficulties in training due to problems like vanishing or exploding gradients.
    \item Number of Heads ($num_{heads}$): More heads allow the model to focus on different parts of the input for each head. This can lead to better performance, but it also increases the computational cost and the risk of overfitting.
    \item Feed Forward Network Dimension ($d_{ff}$): Increasing this value allows the feed-forward network to learn more complex mappings and can lead to better performance, but it also increases the model size and the risk of overfitting.
    \item Learning Rate: If set too high, the model might converge too quickly to a suboptimal solution or might not converge at all. If set too low, the learning process can become too slow. There's usually a sweet spot that needs to be found via experimentation.
    \item Batch Size: Larger batch sizes mean more stable and accurate gradient estimates, at the cost of increased memory usage. However, it's also been found that too large batch sizes can lead to poorer generalization \citep{raffel2020exploring}.
    \item Epochs: Training for more epochs can lead to better performance on the training set, but also increases the risk of overfitting. Early stopping techniques can be used to mitigate this.
    \item Warmup Steps: This hyperparameter is specific to the learning rate scheduler used in transformers. It helps in stabilizing the learning rate during the initial phase of training.
    \item Dropout Rate: Dropout is a regularization technique. A higher dropout rate increases the amount of regularization, which can help prevent overfitting.
    \item Max Sequence Length: This hyperparameter can affect both the computational cost and the kinds of sequences the model can handle. If set too small, important context might be lost.
    \item Parameter Sharing Group Size: Grouped parameter sharing can help reduce overfitting by reducing the number of unique parameters. However, if set too high, the model might not be able to learn complex hierarchical representations effectively.
    \item Optimizer: The choice of optimizer can significantly impact model performance and convergence speed. Adam is a common choice for transformer models due to its efficient handling of sparse gradients and adaptive learning rates.
    \item Activation Function: This function adds non-linearity to the model, enabling it to learn more complex patterns. 'relu' and 'gelu' are common choices in transformer models.
    \item Weight Initialization: Proper initialization can help prevent issues such as vanishing or exploding gradients, leading to faster and more stable convergence.
    \item Gradient Clipping: This is used to prevent exploding gradients, which can cause numerical instability and poor performance \citep{hendrycks2016gaussian}.
    \item Learning Rate Decay: This involves reducing the learning rate as training progresses to enable fine-tuning of the model parameters in the later stages of training.
    \item Regularization Techniques: Apart from Dropout, there are other regularization techniques like L1 and L2 regularization, and also techniques like layer normalization that help to stabilize learning and prevent overfitting.
\end{enumerate}

The appropriate setting of these hyperparameters is task-dependent and usually requires a fair amount of trial and error, or more systematic approaches like grid search or Bayesian optimization, to find the most effective combination. They can significantly affect the time it takes to train your model, the quality of the model, and its ultimate performance.

Table~\ref{param-dmodel} shows the effect of changing the $d_{model}$ hyperparameter on the model's performance:

\begin{table}[!htp]
\centering
\small
\caption{Effects of hyperparameter $d_{model}$}\label{param-dmodel}
\begin{tabular}{p{0.2\linewidth} p{0.39\linewidth} p{0.39\linewidth}}
\toprule
\textbf{Hyperparameter: $d_{model}$} & \textbf{Increase}  & \textbf{Decrease} \\ 
\midrule
Model Complexity & Increases the capacity of the model to learn more complex representations. & Reduces the capacity of the model, potentially making it unable to learn complex patterns. \\
Risk of Overfitting & Increases, as a larger model with more parameters might fit the noise in the training data. & Reduces, as a smaller model has less capacity to fit the noise in the training data. \\
Computational Resources & Requires more computational resources, including memory and processing power. & Requires fewer computational resources. \citep{ba2014deep}\\
Training Time & Likely increases due to the increased number of parameters. & Likely decreases due to fewer parameters to learn. \\
Generalization Ability & Might decrease if the model is too large and starts to overfit. & Might increase up to a point, as a smaller model may generalize better, but if the model is too small, it might underfit the data. \\
\bottomrule
\end{tabular}
\end{table}

It should be noted that these effects are not absolute and may vary depending on the other hyperparameters and the specific task the model is being trained for. The optimal $d_{model}$ value usually requires empirical tuning.

\begin{table}[!htbp]
\centering
\small
\caption{Analysis of the effects of hyperparameter $d_{model}$ on completion tasks}\label{param-anal-lm}
\begin{tabular}{p{0.18\linewidth} p{0.06\linewidth} p{0.06\linewidth} p{0.15\linewidth} p{0.15\linewidth} p{0.12\linewidth} p{0.12\linewidth}}
\toprule
\textbf{Model} & \textbf{$d_{model}$}  & \textbf{$n_{params}$} & \textbf{LAMBADA (acc)} & \textbf{LAMBADA (ppl)} & \textbf{StoryCloze (acc)} & \textbf{HellaSwag (acc)} \\ 
\midrule
GPT-3 Zero-Shot & 12288 & 175B & 76.2 & 3.00 & 83.2 & 78.9   \\ 
GPT-3 One-Shot & 12288 & 175B & 72.5 & 3.35 & 84.7 & 78.1   \\ 
GPT-3 Few-Shot & 12288 & 175B & 86.4 & 1.92 & 87.7 & 79.3   \\ 
GPT-Efficio & 2048 & 950M & 67.1 & 9.2 & 80.5 & 72.6 \\
GPT-Efficio & 1536 & 660M & 58.83 & 11.76 & 58.49 & 68.34 \\
GPT-Efficio & 1024 & 290M & 47.79 & 14.43 & 50.21 & 59.32 \\

\bottomrule
\end{tabular}
\end{table}

Table~\ref{param-anal-lm} demonstrates the GPT-Efficio performance in comparison with GPT-3. 
\begin{figure}
\begin{tikzpicture}
\begin{axis}[
    width=\linewidth,
    height=8cm,
    ybar, ymin=0, ymax=100,
    bar width=.2cm,
    legend style={at={(0.5,-0.15)},
      anchor=north,legend columns=-1},
    ylabel={Accuracy},
    symbolic x coords={GPT-3 Zero-Shot, GPT-3 One-Shot, GPT-3 Few-Shot, GPT-Efficio 2048, GPT-Efficio 1536, GPT-Efficio 1024},
    xtick=data,
    nodes near coords,
    x tick label style={rotate=45,anchor=east},
    legend style={at={(1,1)},
        anchor=south east,legend columns=-1}
    ]
\addplot coordinates {(GPT-3 Zero-Shot,76.2) (GPT-3 One-Shot,72.5) (GPT-3 Few-Shot,86.4) (GPT-Efficio 2048,67.1) (GPT-Efficio 1536,58.83) (GPT-Efficio 1024,47.79)};
\addplot coordinates {(GPT-3 Zero-Shot,83.2) (GPT-3 One-Shot,84.7) (GPT-3 Few-Shot,87.7) (GPT-Efficio 2048,80.5) (GPT-Efficio 1536,58.49) (GPT-Efficio 1024,50.21)};
\addplot coordinates {(GPT-3 Zero-Shot,78.9) (GPT-3 One-Shot,78.1) (GPT-3 Few-Shot,79.3) (GPT-Efficio 2048,72.6) (GPT-Efficio 1536,68.34) (GPT-Efficio 1024,59.32)};
\legend{LAMBADA,StoryCloze,HellaSwag}
\end{axis}
\end{tikzpicture}
\caption{Analysis of the effects of hyperparameter $d_{model}$ on completion tasks}
\end{figure}
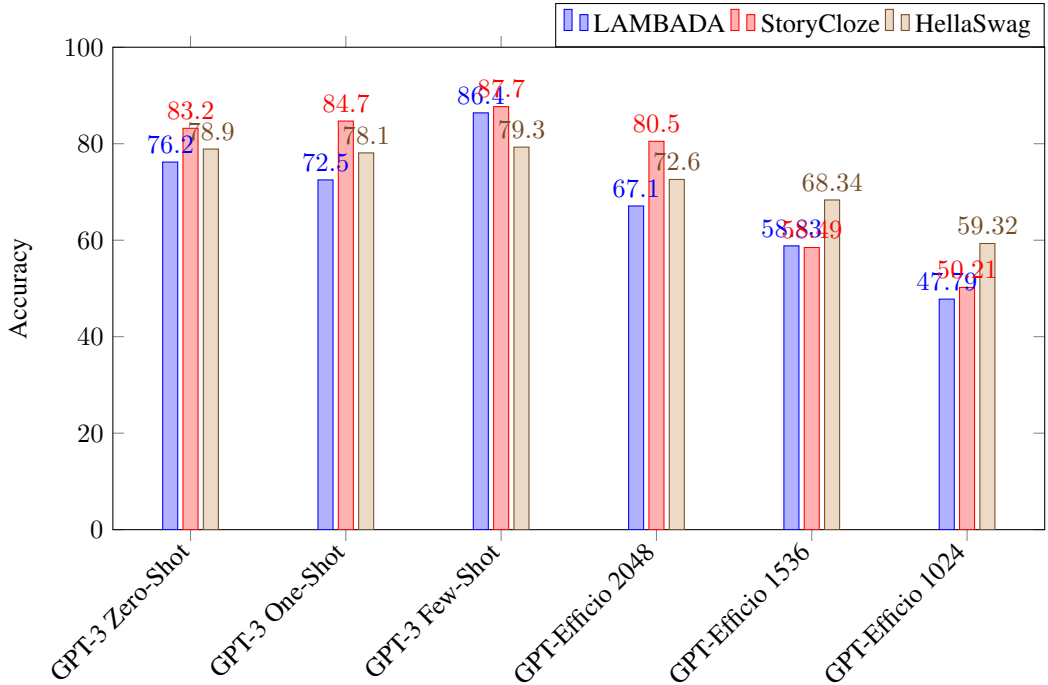

\begin{table}[!htbp]
\centering
\small
\caption{Analysis of the effects of hyperparameter $d_{model}$ on QA tasks}\label{param-anal-qa}
\begin{tabular}{p{0.2\linewidth} p{0.1\linewidth} p{0.1\linewidth} p{0.15\linewidth} p{0.15\linewidth} p{0.12\linewidth}}
\toprule
\textbf{Model} & \textbf{$d_{model}$} & \textbf{$n_{params}$} & \textbf{NQ} & \textbf{WebQ} & \textbf{TriviaQA}\\ 
\midrule
GPT-3 Zero-Shot & 12288 & 175B & 14.6 & 14.4 & 64.3   \\ 
GPT-3 One-Shot & 12288 & 175B & 23.0 & 25.3 & 68.0   \\ 
GPT-3 Few-Shot & 12288 & 175B & 29.9 & 41.5 & 71.2   \\ 
GPT-Efficio & 2048 & 950M & 27.5 & 40.6 & 69.2 \\
GPT-Efficio & 1536 & 660M & 19.32 & 31.86 & 58.49 \\
GPT-Efficio & 1024 & 290M & 15.28 & 27.67 & 50.21 \\
\bottomrule
\end{tabular}
\end{table}

Table~\ref{param-anal-qa} shows the GPT-Efficio performance in comparison with GPT-3. 

\begin{figure}[htbp]
\centering
\begin{tikzpicture}
\begin{axis}[
    width=\linewidth,
    height=7cm,
    ybar,
    ymin=0,
    ylabel={Accuracy},
    xtick=data,
    xticklabels={GPT-3 Zero-Shot, GPT-3 One-Shot, GPT-3 Few-Shot, GPT-Efficio (2048), GPT-Efficio (1536), GPT-Efficio (1024)},
    nodes near coords,
    nodes near coords align={vertical},
    x tick label style={rotate=10,anchor=east},
    legend style={at={(1,1)},
    anchor=south east,legend columns=-1}
    ]
\addplot coordinates {(1,14.6) (2,23.0) (3,29.9) (4,27.5) (5,19.32) (6,15.28)};
\addplot coordinates {(1,14.4) (2,25.3) (3,41.5) (4,40.6) (5,31.86) (6,27.67)};
\addplot coordinates {(1,64.3) (2,68.0) (3,71.2) (4,69.2) (5,58.49) (6,50.21)};
\legend{NQ, WebQ, TriviaQA}
\end{axis}
\end{tikzpicture}
\caption{Analysis of the effects of hyperparameter $d_{model}$ on QA tasks}
\label{fig:param-anal-qa}
\end{figure}
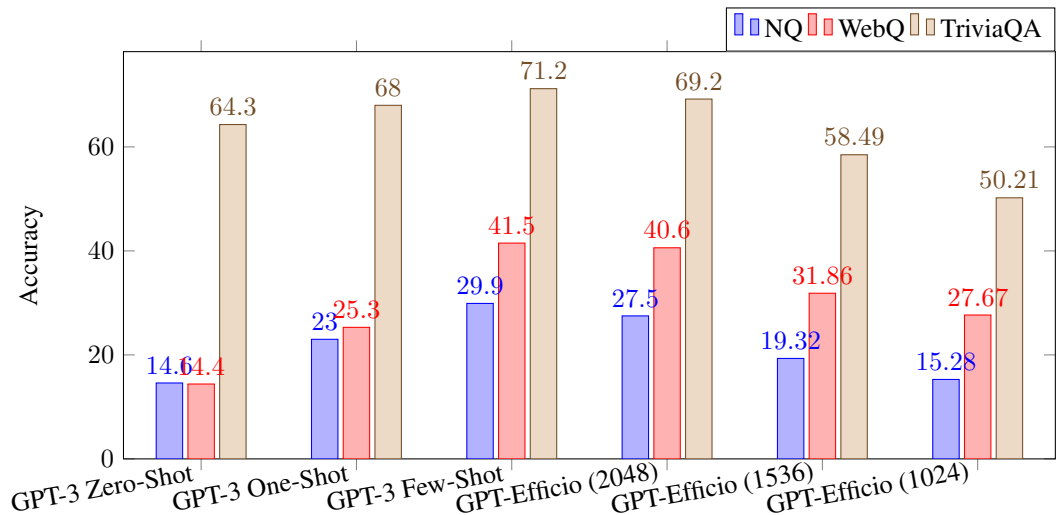

\section{Limitations}

Parameter sharing techniques in transformer-based models are used to mitigate issues related to the high computational cost and complexity that accompany these models. However, despite the apparent benefits, there are several limitations associated with the application of these techniques:

\begin{itemize}
    \item Limited Model Expressivity: One of the most pronounced drawbacks of parameter sharing is the potential reduction in model expressivity. By using the same parameters across different layers or parts of the model, we inherently restrict the ability of the model to learn and represent diverse features and relationships. This could lead to a lack of depth in the learned representations, potentially impacting the model's performance on complex tasks that require the learning of diverse and intricate patterns.

    \item Sub-optimal Solutions: In certain cases, the optimal solution might require unique parameters in different parts of the model. By enforcing parameter sharing, we might be pushing the model towards sub-optimal solutions. This could lead to poor model performance, especially in tasks where the learning of unique representations at different parts or depths of the model is critical.

    \item Task and Data Dependence: The effectiveness of parameter sharing techniques is not universal and depends heavily on the specific task and data at hand. For instance, parameter sharing might work well for some tasks, such as text generation or sequence-to-sequence prediction, where recurrent patterns exist. However, for tasks where the requirement of unique representations across the model is high, parameter sharing could negatively impact performance.

    \item Risk of Over-simplification: While reducing complexity is one of the motivations behind parameter sharing, there's a risk of oversimplifying the model to the extent that it fails to capture the necessary complexity of the data. This can lead to underfitting, where the model is too simple to capture the underlying structure of the data, leading to poor performance.

    \item Difficulty in Optimization: Sharing parameters can make the optimization landscape more complex. When parameters are tied, a change in one location means a change in another, potentially leading to more difficult optimization and slower convergence.
\end{itemize}
While parameter sharing offers valuable advantages, it's not a one-size-fits-all solution. Consideration must be given to the specific nature of the task, the complexity of the data, and the capacity of the model to ensure that parameter sharing benefits outweigh the potential limitations.

\section{Future Work}
Given the demonstrated effectiveness as well as the limitations of parameter sharing techniques in transformer-based models like GPT, future work in this domain could be directed towards several interesting and promising avenues:

\begin{itemize}
    \item Adaptive Parameter Sharing: Developing methods for adaptive parameter sharing could be an exciting line of research. These methods would adjust the extent of parameter sharing based on the specific task or data. This could involve learning which parts of the model should share parameters and to what extent, potentially through the use of techniques like reinforcement learning or meta-learning.

    \item Hybrid Parameter Sharing Schemes: Exploring hybrid parameter sharing schemes that combine different types of parameter sharing could be beneficial. For instance, some layers could share parameters while others could have unique parameters, or different types of parameter sharing could be used for different parts of the model.

    \item Task-specific Parameter Sharing Strategies: Investigating task-specific parameter sharing strategies could lead to improvements in performance. For example, different parameter sharing techniques could be developed and evaluated for tasks like text generation, translation, summarization, or question-answering.

    \item Learning Shared and Unique Representations: Research could be directed towards models that can learn both shared and unique representations simultaneously. This could involve architectures that have a mix of shared and unique parameters, allowing them to benefit from the generalization of parameter sharing while also being able to learn task or data-specific features.

    \item Theoretical Analysis of Parameter Sharing: Further theoretical analysis of parameter sharing could lead to a better understanding of its impacts on model capacity, generalization, optimization, and other aspects of model performance. This could involve both analysis of existing methods and the development of new theoretical models or frameworks.

    \item Improved Optimization Techniques: Given the potential difficulties in optimizing models with shared parameters, developing improved optimization techniques for such models could be beneficial. This could include new training algorithms or regularization techniques designed specifically for models with parameter sharing.

    \item Evaluation Across Diverse Tasks and Datasets: Lastly, systematic evaluation of parameter sharing techniques across a wider range of tasks and datasets could provide valuable insights into when and why these techniques are effective or ineffective. This could guide the development of more robust and adaptable parameter sharing methods.
\end{itemize}

By pursuing these directions, future work could overcome some of the limitations of current parameter sharing techniques and lead to more efficient and effective transformer-based models.

\section{Conclusion}
In this approach, we introduced a new Large Language Model, GPT-Efficio, that implements two techniques we call "Embedding layer factorization" and "Parameter sharing across layers". The application of parameter sharing strategies in transformer-based models such as the Generative Pretrained Transformer (GPT) can produce a myriad of potential outcomes, each with significant implications for the model's performance, computational efficiency, and generalization capabilities.

In particular, these techniques offer a way to address several challenges endemic to machine learning, such as reducing model complexity, speeding up training processes, improving generalization, and mitigating overfitting risks. These potential benefits highlight the value of parameter sharing, especially in situations where computational resources, memory, or time are constraining factors, or where the available training data is limited.

While parameter sharing techniques provide promising avenues for optimizing transformer-based models, the decision to implement such strategies should be made judiciously, guided by the specific requirements of the task and empirical evidence. Careful considerations must be made to balance the trade-off between model complexity and expressivity, and extensive experimentation and validation are crucial to ascertain the most effective application of these methods.

\bibliographystyle{plainnat}
\bibliography{main}

\end{document}